\documentclass{article}
\usepackage{spconf}
\usepackage{amsmath,graphicx}

\usepackage{amssymb}
\usepackage{algorithm,algcompatible}
\usepackage{multirow}
\usepackage{rotating}
\usepackage{subfig}
\usepackage{todonotes}
\usepackage{authblk}
\usepackage{url}
\usepackage{hyperref}

\newtheorem{definition}{Definition}

\newcommand\blfootnote[1]{%
  \begingroup
  \renewcommand\thefootnote{}\footnote{#1}%
  \addtocounter{footnote}{-1}%
  \endgroup
}

\usepackage{caption}

\title{What does the user want? Information Gain for Hierarchical Dialogue Policy Optimisation}

\name{C. Geishauser$^1$, S. Hu$^2$, H. Lin$^1$, N. Lubis$^1$, M. Heck$^1$, S. Feng$^1$, C. van Niekerk$^1$, M. Gašić$^1$}

\address{{$^1$\textit{Heinrich Heine University Düsseldorf}, Germany}\\ {$^2$\textit{Department of Computer Science and Technology, University of Cambridge}, UK}\\ \texttt{\small \{geishaus, linh, lubis, heckmi, shutong.feng, niekerk, gasic\}@hhu.de, \small sh2091@cam.ac.uk}}

\begin{document}
\maketitle
\begin{abstract}
The dialogue management component of a task-oriented dialogue system is typically optimised via reinforcement learning (RL). Optimisation via RL is highly susceptible to sample inefficiency and instability. The hierarchical approach called Feudal Dialogue Management takes a step towards more efficient learning by decomposing the action space. However, it still suffers from instability due to the reward only being provided at the end of the dialogue. We propose the usage of an intrinsic reward based on information gain to address this issue. Our proposed reward favours actions that resolve uncertainty or query the user whenever necessary. It enables the policy to learn how to retrieve the users' needs efficiently, which is an integral aspect in every task-oriented conversation. Our algorithm, which we call FeudalGain, achieves state-of-the-art results in most environments of the PyDial framework, outperforming much more complex approaches. We confirm the sample efficiency and stability of our algorithm through experiments in simulation and a human trial.
\end{abstract}
\section{Introduction}
\label{sec:intro}
\blfootnote{To be published in {\it \ ASRU2021, December 13-17, 2021, Cartagena, Colombia}. \copyright\ IEEE 2021. Personal use of this material is permitted. Permission from IEEE must be obtained for all other uses, in any current or future media, including reprinting/republishing this material for advertising or promotional purposes, creating new collective works, for resale or redistribution to servers or lists, or reuse of any copyrighted component of this work in other works. }

Task-oriented dialogue  systems are characterised by an underlying task or a goal that needs to be achieved during the conversation in order to help a user, such as managing a schedule or finding and booking a restaurant. For that, a spoken dialogue system needs two key abilities: maintaining the current state of the dialogue (\textit{tracking}) and foreseeing how its actions will impact the conversation (\textit{planning}). Modular dialogue systems therefore have a tracking component to maintain information about the dialogue belief state, and a planning component that models the underlying policy, i.e. the selection of actions~\cite{Levin97astochastic, probabilistic-roy, POMDP-Williams, zhang2020recent}. The dialogue belief state defines a probability distribution over states that includes information about user preferences, for instance that a user wants a cheap, Italian restaurant, with the distribution encoding different levels of uncertainty.

To deal with planning, current state-of-the-art dialogue systems \cite{strac, diluted, lava} optimise the policy via some form of reinforcement learning (RL) \cite{Sutton}. However, dialogue policy optimisation using RL is often sample inefficient and unstable, which is exacerbated by the sparse reward typically given in task-oriented dialogue systems \cite{acer1, challengesRL2, challengesRL1, zhang2020recent}. 
To tackle the problem of sample inefficiency, hierarchical reinforcement learning has been proposed that subdivides the task temporally or spatially \cite{optionssutton, hinton93}, thereby reducing complexity of the task and accelerating learning. 

For spoken dialogue systems that help users accomplish any kind of tasks, it is important to understand what the actual user's goal is by asking appropriate and targeted questions. Acquiring information is the first important building block of a conversation. Due to its significance, it is reasonable to learn a dedicated policy $\pi_i$ that deals with this sub-task. This has been proposed in the hierarchical approach called Feudal Dialogue Management \cite{feudal1}, where the extrinsic reward is used for optimising $\pi_i$. However, we argue and show that the extrinsic reward may provide misleading feedback signal for $\pi_i$ that leads to unstable and less efficient learning. 

Instead, we equip the policy with an intrinsic reward based on information gain that measures the change in probability distributions between consecutive turns. The dense reward signal encourages the policy to produce actions that resolve uncertainty and to query the user in cases where it is necessary. The policy $\pi_i$ together with our proposed reward explicitly models how a system can learn to obtain information about the dialogue partner, which is an integral aspect in every conversation.

We conduct our experiments using the PyDial benchmarking environment \cite{pydial}. 
Our algorithm \emph{FeudalGain} achieves state-of-the-art results in terms of sample efficiency and final performance in 14 out of 18 environments. We confirm the effectiveness of our method in a human trial, where our system directly interacts with humans.

\section{Related Work}

Information gain (also known as mutual information) measures the amount of information obtained about one random variable through observing another random variable. Information gain has already been used as feature as well as reward signal for reinforcement learning. In \cite{padmakumar2020dialog}, a dialogue policy for clarification is trained using information gain as a policy feature. Information gain has also been used to build decision trees for dialogue systems \cite{ig1, dt}. 

Different to the extrinsic reward that is produced by the environment, intrinsic reward is produced by the RL agent itself. The purpose of intrinsic reward is to additionally guide the learning of the agent. Curiosity-driven learning, which encourages exploration of the state-action space by learning more about the environment dynamics, has been interpreted as information gain \cite{vime, bbq}. In contrast to curiosity that will decrease the more the agent learns about the environment, the information gain we use will always be provided. In \cite{intrinsichrl},  the policy is divided into an ``explorer'' and ``exploiter'', using an intrinsic reward signal for the explorer that is different from the extrinsic reward for the exploiter. While \cite{intrinsichrl} focus on exploration, we define an intrinsic reward to foster fast learning of an information seeking policy. Note that even when the space is fully explored, one still needs to gather sufficient information about user needs. 

``Answerer in Questioner’s Mind'' \cite{ig2} selects the question which maximises the information gain of a target class and an answer given the question in a goal-oriented visual dialogue task. In \cite{whatask, goldseeker}, information gain is used as a reward in goal-oriented visual dialogues by leveraging a responder model or guesser. 
The belief tracker in dialogue can be interpreted as a guesser that needs to guess the correct value for each slot in the belief state, where the policy can ask questions. Different from \cite{whatask, goldseeker}, we use information gain in a hierarchical setting and provide it only to a sub-policy as (intrinsic) reward. 

In \cite{feudal1}, there is a dedicated sub-policy for information gathering as in our case.  It is however only optimised with the extrinsic reward. We introduce an intrinsic reward based on information gain in the hierarchical setting and thus enable fast learning of the user's needs, an integral ability that is often neglected in task-oriented dialogue systems.


\section{Background}

\subsection{Dialogue Policy Optimisation via RL}
The formal framework in RL is given by a Markov decision process $\mathcal{M}=\{\mathcal{B}, \mathcal{A}, r, p, p_0, \gamma\}$. Here, $\mathcal{B}$ denotes the (continuous) belief state space, $\mathcal{A}$ is the action space, $r$ is the reward function,  $p(b^\prime | b, a)$ models the probability of transitioning to state $b^\prime$ after executing action $a$ in state $b$, and $p_0(b)$ is the probability of starting in state $b$. The discount factor $\gamma \in [0,1]$ trades off the importance of immediate and future rewards. At time step $t$, the agent observes a state $b_t$, chooses an action $a_t$ according to a policy $\pi(a|b_t)$, transitions to a new state $b_{t+1}$ and observes a reward signal $r_t = r(b_t, a_t, b_{t+1}) \in \mathbb{R}$. The goal of the agent is to maximise the discounted return $R_t = \sum_{i \geq 0} \gamma^{i} r_{t+i}$ in expectation. 

Value-based RL methods \cite{Sutton} optimise the policy by maximising the Q-values $Q^\pi(b_t, a_t) = \mathbb{E}_{b_{t+1:\infty}, a_{t+1:\infty}} [R_t | b_t, a_t]$ for every state-action pair $(b_t, a_t) \in \mathcal{B} \times \mathcal{A}$. One example here is Dueling deep Q-networks (DDQN) \cite{ddqn} that optimises a parameterised Q-network using tuples $({b_t, a_t, r_t, b_{t+1}})$ and a target network for a stochastic gradient step. Policy gradient methods on the other hand parameterise the policy directly and aim at maximising $\mathbb{E}_{b_0}[R_0]$. Actor-critic algorithms \cite{Sutton} build upon policy gradient methods and approximate $Q^\pi$ with a function approximator, also called critic. The ACER \cite{acer1} algorithm is one such instance that uses the whole trajectory in order to do an update step for its critic.  

To foster exploration during learning, noisy networks \cite{noisynetwork} inject noise into the neural network by substituting each weight $w$ in the neural network by $\mu + \sigma \cdot \epsilon$, where $\mu$ and $\sigma$ are weights and $\epsilon$ is a noise random variable. 

We view dialogue as a sequence of turns between a user and a dialogue system. The belief state in dialogue typically includes a probability distribution over values for every slot in the ontology, which expresses how likely it is that the user wants a specific value such as ``Italian'' food or ``expensive'' price-range. In each turn of the dialogue, for a given belief state, the system decides which action to take in order to successfully complete the user's goal. This sequential decision-making task can be optimised with RL algorithms. A range of RL algorithms have already been applied to dialogue management, including ACER~\cite{acer2}.


\begin{table*}[t!]
  \begin{center}
  \resizebox{1.8\columnwidth}{!}{%
    \begin{tabular}{l|c|c|c|c|c|c} 
       & Utterance & $p_s$ for $s=\text{\emph{pricerange}}$ & $r_e$ & $r_i$ & $\pi_i$ & $\pi_g$\\
      \hline
      User & I need a restaurant. & [0,0,0,0,1] &  & & & \\
      \hline
      Sys ($\pi_i$) & What pricerange do you like? & [0,0,0,0,1] & -1.0 & 1.0 & \emph{request-pricerange} & \emph{pass}\\
      \hline
      User & Something cheap please. $\rightarrow$ & [0.5,0.3,0.2,0,0] & & & &\\
      \hline
      Sys ($\pi_i$) & Can I confirm that you mean cheap? & [0.5,0.3,0.2,0,0] & -1.0 & 0.22 & \emph{confirm-pricerange} & \emph{pass}\\
      \hline
      User & Yes. I need the address. $\rightarrow$ & [0.95,0.05,0,0,0] & & & &\\
      \hline
      Sys ($\pi_g$) & Goodbye. & [0.95,0.05,0,0,0] & 0.0 & 0.0 & \emph{pass} & \emph{bye}\\
    \end{tabular}
    }
    \caption{Example dialogue in the Cambridge restaurant domain. $p_s$ denotes the probability distribution over values [\emph{cheap}, \emph{moderate}, \emph{expensive}, \emph{dontcare}, \emph{none}] for slot \emph{pricerange} given in the belief state. $r_e$ and $r_i$ denote extrinsic and intrinsic reward given to $\pi_i$ during the conversation. $\pi_i$ tries to retrieve information and resolve uncertainty, yet the extrinsic reward gives no guidance at all since $\pi_g$ ends the dialogue too early. In contrast, our proposed reward $r_i$ correctly rewards the behaviour of $\pi_i$. Information gain $r_i$ given in Definition \ref{rewarddefinition} is computed by Jensen-Shannon divergence between consecutive value distributions.}
        \label{faileddialogue}
    
  \end{center}
\end{table*}

\subsection{Feudal Dialogue Management}
Feudal Dialogue Management \cite{feudal1, feudal2} is a hierarchical approach for dialogue policy learning that divides the action space $\mathcal{A}$ into two subsets $\mathcal{A}_i$ and $\mathcal{A}_g$. The purpose of actions in $\mathcal{A}_i$ is to obtain more information from the user by \emph{confirming}, \emph{requesting} or \emph{selecting} the value of a slot. The second action set $\mathcal{A}_g$ comprises all other actions, such as general actions like \emph{goodbye} or informing about requested values. Moreover, a master action space $\mathcal{A}_m = \{a_i, a_g\}$ for choosing between actions in $\mathcal{A}_i$ and $\mathcal{A}_g$ is defined. In order to produce an action, a master policy $\pi_m$ first selects an action from $\mathcal{A}_m$, after which the associated policy $\pi_i$ or $\pi_g$ corresponding to $\mathcal{A}_i$ and $\mathcal{A}_g$ is consulted for the final action selection. An additional \emph{pass} action is added to $\mathcal{A}_i$ and $\mathcal{A}_g$, which is taken whenever the other sub-policy is executed. The policy $\pi_i$ is optimised using a value-based method where the Q-values are produced for every slot $s \in \{s_1,...,s_n\}$ independently using associated policies $\pi_s$. The input to each $\pi_s$ is the belief state including the value distribution of $s$. The parameters of the Q-functions are shared among the slots. 

To optimise each of the policies, the external reward $r_e$ provided by the environment is used. The reward is $-1$ in each turn to enforce more efficient dialogues and $0$ or $20$ in the very last turn for failure or success of the dialogue.

We note that adding the \emph{pass} action for $\pi_i$ is very important. The reward for a tuple $(b_t, a_t, r_t, b_{t+1})$ with $a_t \neq \text{\emph{pass}}$ that we use for updating $\pi_i$ will always be $-1$ or $0$ since success can be only achieved if information is provided (which only $\pi_g$ can do). We hence need to update the policy using tuples where the \emph{pass} action was taken, i.e. reward the policy for doing nothing. We empirically verify the necessity for the \emph{pass} action in Section \ref{ablationstudy}.

In the following, we will work with the FeudalACER algorithm \cite{feudal2} as our baseline and abbreviate it as Feudal. Feudal uses ACER for policies $\pi_m$ and $\pi_g$ and DDQN for $\pi_i$.

\section{Information Gain in Policy Learning}

\subsection{Drawbacks of Extrinsic Reward in Feudal}
Recall that $\pi_i$ merely outputs actions for obtaining information about the user preferences. As this is not enough to complete a task, the policy $\pi_g$ mainly determines dialogue success or failure. While it is reasonable to provide $\pi_g$ and $\pi_m$ with external reward, it is less obvious for $\pi_i$. The behaviour of $\pi_i$ can lead either to reinforcement or suppression if $\pi_g$ misbehaves. As an illustrative example, Table \ref{faileddialogue} shows a dialogue where $\pi_i$ acted correctly but does not obtain any positive feedback from $r_e$ due to dialogue failure in the end. 

\subsection{Information Gain}

How can we fairly reward $\pi_i$ then? We propose the usage of an intrinsic reward for $\pi_i$ based on information gain similar to \cite{whatask}. The idea is that if we take an action to query information about a certain slot (e.g. \emph{request-area}) leading to a change in the value distribution for that slot, new information has been gathered and that behaviour should be reinforced. Formally, we define the intrinsic reward $r_i$ as follows.

\begin{definition}\label{rewarddefinition}
Let $(b, a, b^\prime) \in \mathcal{B} \times \mathcal{A}_i \times \mathcal{B}$ be a tuple of state, action and next state where $a$ includes slot $s$. Let $p_s$ and $p^\prime_s$ be the probability distributions over values for $s$ in b and $b^\prime$, respectively. Let $d$ be a distance function between probability distributions. We define
\begin{equation*}
    r_i(b, a, b^\prime) := d(p_s, p^\prime_s)
\end{equation*}
as the information gain (IG) when executing action $a$ in state $b$ and observing $b^\prime$. 

\end{definition}
This reward encodes the goal of $\pi_i$ by reinforcing actions that gather new information or resolve uncertainty. It separates learning of $\pi_i$ from the behaviour of $\pi_g$ and independently models how a system can learn to obtain information about the user's needs. Moreover, the reward guides the policy at every step in contrast to the sparse reward that first has to be back-propagated. Due to the immediate feedback, the additional \emph{pass} action becomes obsolete for $\pi_i$ and we do not need to update with tuples $(b_t, \text{\emph{pass}}, r_t, b_{t+1})$ anymore. The policy can now quickly learn how to obtain the user preferences, which is the first important step towards a successful dialogue. An example for computing $r_i$ together with the chosen actions is depicted in Table ~\ref{faileddialogue}. In contrast to the external reward, information gain reinforces the desired behaviour of $\pi_i$ even though the dialogue failed. Since the probability distribution is part of the input to the policy, $\pi_i$ can easily build the relation between the state and the reward. We note that this reward is only defined and should be only used for actions that seek to obtain information of the user. Otherwise it might happen that the user pro-actively provides information and an unrelated action gets rewarded. As a result, the reward can be applied to all scenarios where obtaining information from a user is important. This is especially the case for task-oriented dialogue systems where our focus lies, but also holds in many more scenarios (conducting an interview, getting to know a person in chit-chat) as dialogue is generally an exchange of information.

We remark that the usage of our reward is not restricted to hierarchical RL and it can also be used as an additional signal to the external reward. We also emphasise that our dense reward aids the policy in learning to reach the main goal, i.e. task completion. It thus differs from rewards based on curiosity or surprise \cite{vime, pathak} that aim for enhanced exploration (which decreases as the agent learns more about the environment). Our reward can be used in tandem with other rewards, in particular the ones for exploration.


\begin{center}
\begin{figure}[h]
\includegraphics[trim=0.0cm 0cm 0.0cm 1cm, width=0.49\textwidth]{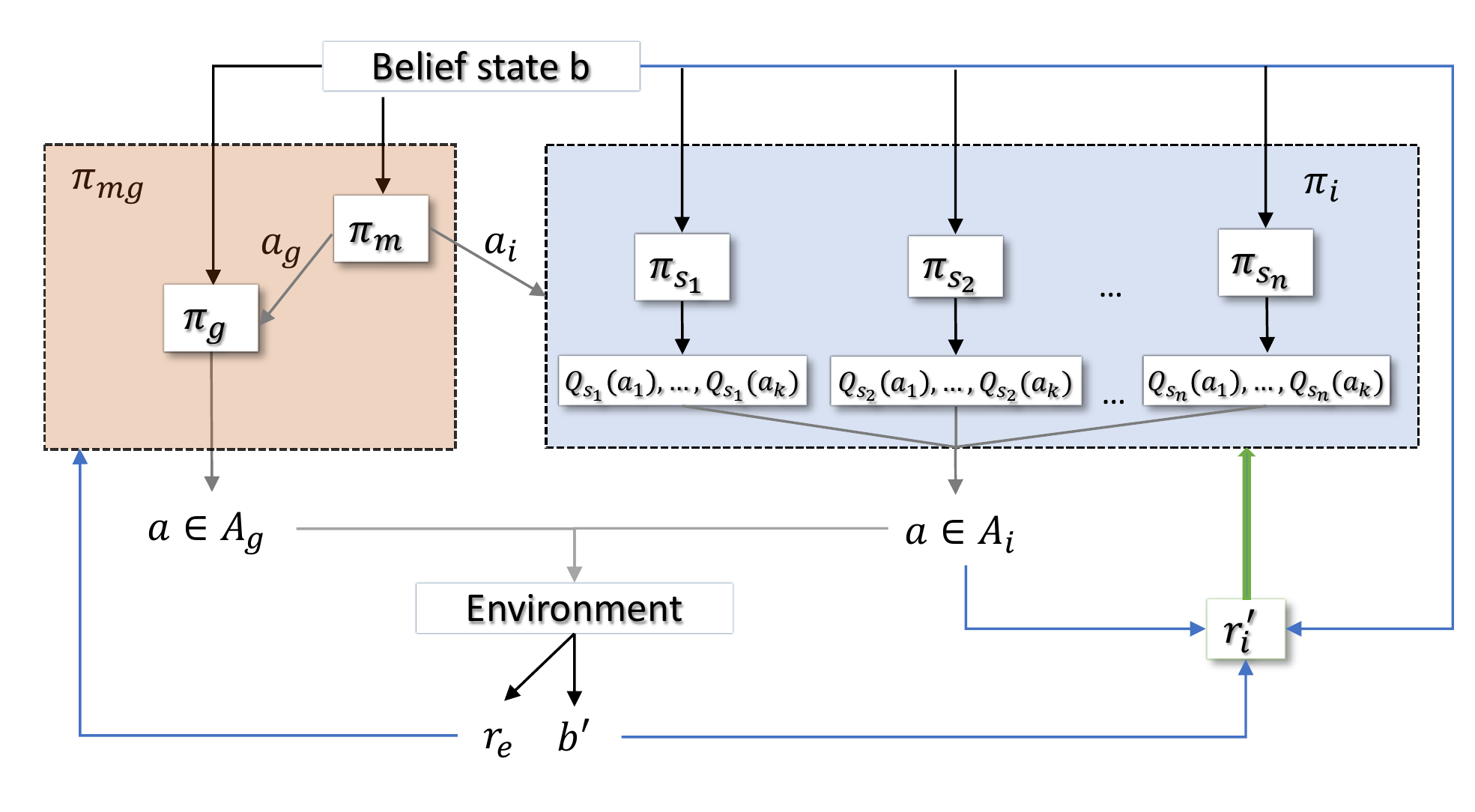}
\caption{The architecture of our FeudalGain algorithm together with the action selection process and reward computation of $r_i^\prime$. FeudalGain uses the intrinsic reward $r_i^\prime$ as given in Equation (\ref{intrinsicreward}) instead of the external reward $r_e$ for optimising $\pi_i$. Also different from Feudal, FeudalGain merges the policies $\pi_m$ and $\pi_g$ into a single policy $\pi_{mg}$ with action space $\mathcal{A}_g \cup \{a_i\}$.}
\label{feudalarchitecture}
\end{figure}
\end{center}

\vspace{-1.4cm}
\subsection{FeudalGain}

After introducing our intrinsic reward based on information gain, we now present our FeudalGain algorithm. Most importantly, we substitute the extrinsic reward for our proposed information gain to optimise the policy $\pi_i$. 

We choose Jensen-Shannon divergence (JS) as our distance function $d$, since it is bounded between $0$ and $1$, symmetric and defined everywhere, in contrast to Kullback-Leibler-divergence (KL) \cite{kl}. JS for two probability distributions $p$ and $q$ is defined as
\begin{align*}
    \text{JS}(p, q) &= \frac{1}{2} (\text{KL}[p || m ] + \text{KL}[q || m]), \\
    m &= \frac{1}{2} (p + q)
\end{align*}

In our experiments, we found that rewarding behaviour if information gain exceeds a certain threshold $\delta$ works better than directly using $r_i$. We henceforth work with the reward
\begin{equation} \label{intrinsicreward}
       r_i^\prime(b, a, b^\prime) = 
\begin{cases}
    1, &\text{if } r_i(b, a, b^\prime) \geq \delta\\
    -1,               &\text{otherwise}
\end{cases} 
\end{equation}
with variables as in Definition \ref{rewarddefinition}. 

ACER uses the full trajectory to update its critic, which is why $\pi_g$ needs to take an action in every turn. Feudal solves this by $\pi_g$ taking the \emph{pass} action whenever $\pi_i$ takes an action as shown in Table \ref{faileddialogue}. However, we can avoid that additional action by merging the policies $\pi_m$ and $\pi_{g}$ into a single policy $\pi_{mg}$ with action space $\mathcal{A}_g \cup \{a_i\}$, which we employ in our full algorithm.

Our final algorithm FeudalGain thus uses $\pi_{mg}$ and $\pi_i$ together with intrinsic reward $r_i^\prime$ for $\pi_i$. Our full algorithm is depicted in Figure \ref{feudalarchitecture}.

\section{Experimental Setup} \label{experiments}

We implement FeudalGain\footnote{Our code will be released at \url{https://pydial.cs.hhu.de}.} in the PyDial toolkit \cite{pydialtk}. The performance is evaluated using the PyDial benchmarking environments \cite{pydial} comprising 18 settings, which are distinguished by domain and different semantic error rates, action masks and user simulator configurations. Unlike other publicly available dialogue toolkits, PyDial uses a belief tracker that outputs probability distributions rather than binary states, which enables more expressive distribution comparisons.


Instead of the $\epsilon$-greedy approach for exploration that was used for Feudal, we use noisy networks similar to \cite{strac}.
The threshold $\delta$ for all our experiments in Section \ref{feudalgainexp} is set to $0.2$. The FeudalGain policies $\pi_{mg}$ and $\pi_i$ are trained with ACER and DDQN, respectively. For each environment, the algorithms are trained on 4000 dialogues with 10 different seeds. After every 200 training dialogues, the algorithms are evaluated on 500 dialogues. The average reward that is shown is always the extrinsic reward $r_e$. The dialogues for our human trial are collected using DialCrowd \cite{dialcrowd}. The simulated user experiments are done on semantic level using the default focus belief tracker, while the user trial is performed on text-level using an additional template based natural language generation module.

We compare FeudalGain to Feudal \cite{feudal2} and the current state-of-the-art algorithm STRAC \cite{strac}. STRAC uses a hierarchical decision-making model for policy optimisation with implicit policy decomposition and noisy networks for exploration. STRAC offers two different modes: a single-domain version STRAC-S that is trained and evaluated on a single domain and STRAC-M that is trained on three different domains but evaluated only on a single domain. STRAC-M trains on three times as many dialogues as STRAC-S and FeudalGain. 
We do not compare to the work of \cite{diluted} as they use a hand-coded expert during training. For completeness, we also add the performance of a hand-coded policy (HDC), which is already implemented in PyDial.

\section{Results}

\subsection{Results on FeudalGain} \label{feudalgainexp}

We compare FeudalGain to STRAC-S in terms of sample efficiency and final performance. Table \ref{tab:sotatable} shows success rate and average reward after 400 and 4000 dialogues.

FeudalGain has higher sample efficiency than STRAC-S in almost all settings and is comparable to STRAC-M although STRAC-M uses three times as many dialogues. This can be attributed to the immediate reward provided by our information gain that correctly guides $\pi_i$ in every turn. 

Similar conclusions can be drawn after 4000 dialogues, where FeudalGain is even able to outperform STRAC-M. Information gain hence not only helps in securing more sample efficient learning but also for achieving high final performance. FeudalGain excels in difficult environments, namely 2 and 4, where unreasonable actions are not masked. FeudalGain also performs very well in environment 6 that exhibits a very high noise level of $30 \%$, which shows that information gain is robust to high error rates. Final performance is slightly worse in environment 5, where an ``unfriendly'' user simulator is used. The results show that policy optimisation can significantly benefit from our reward based on information gain.

\begin{figure}[t!]
    \centering
    \includegraphics[trim=0.0cm 0cm 0.0cm 1.25cm, width=0.5\textwidth]{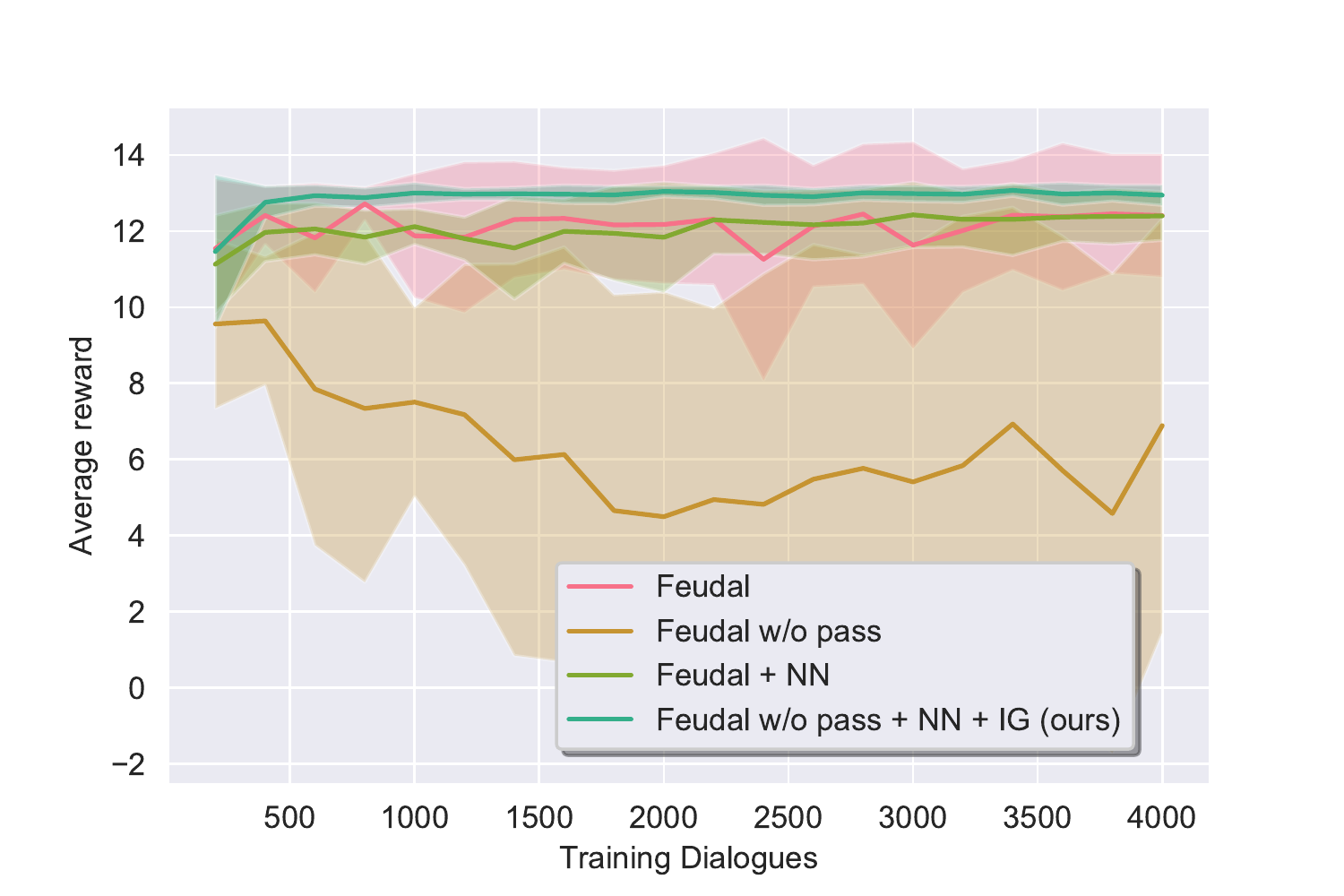}
    \caption{Ablation study for FeudalGain. ``W/o pass'' ablates the additional \emph{pass} action for $\pi_i$. NN denotes noisy networks. IG denotes our proposed information gain addition.}
    \label{fig:feudalresult}
    
\end{figure}

\begin{figure}[t!]
    \centering
    \includegraphics[trim=0.0cm 0cm 0.0cm 0.8cm, width=0.5\textwidth]{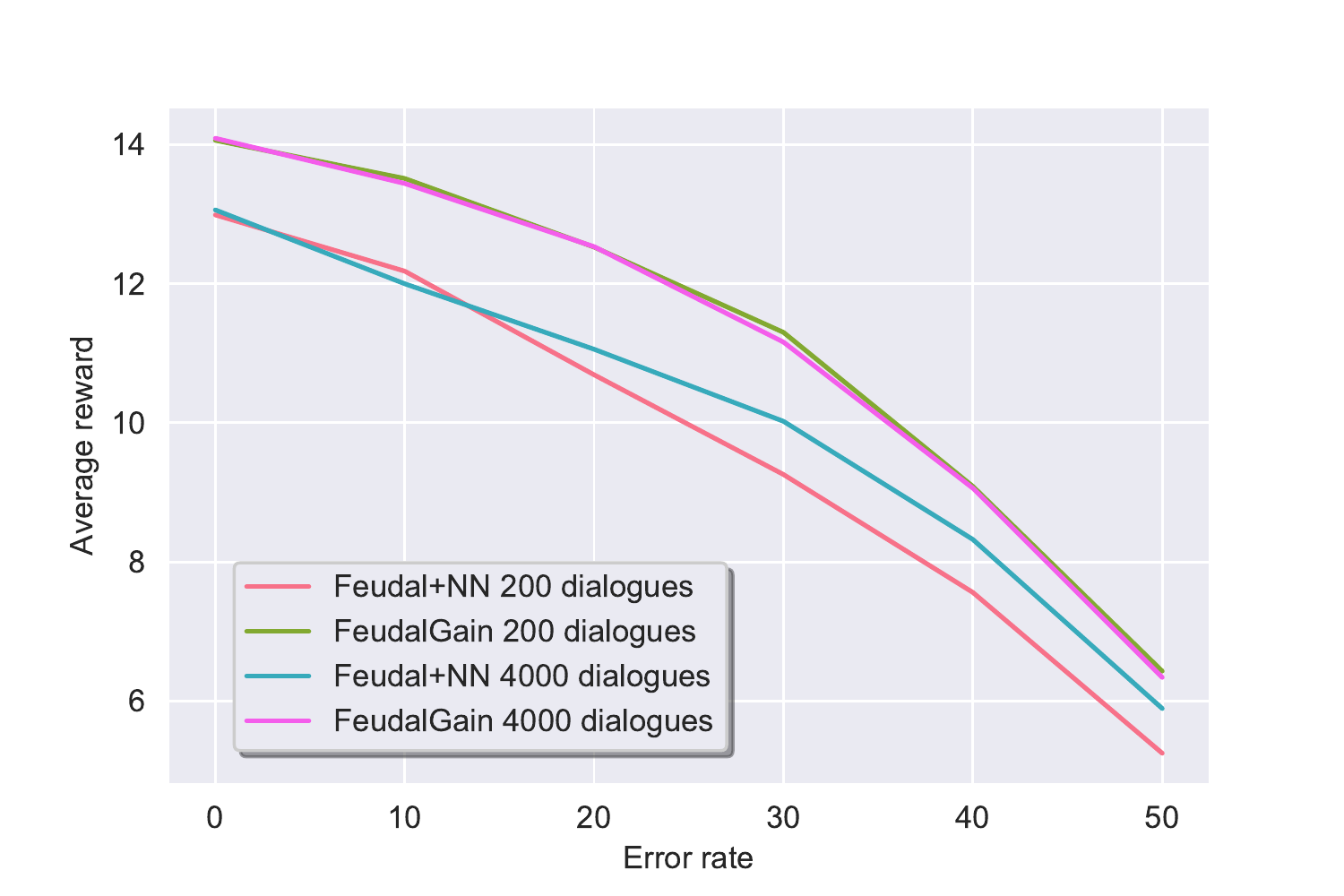}
    \caption{Robustness test of FeudalGain against Feudal+NN for increasing semantic error rate. We use policies that trained for 200 dialogues and 4000 dialogues averaged over 15 seeds.}
    \label{fig:noisetest}
    
\end{figure}

\begin{table}[t!]
\begin{center}
\resizebox{0.95\columnwidth}{!}{%
\begin{tabular}{cccccc|cc|cc}
    \hline
    \multicolumn{2}{c}{} &\multicolumn{2}{c}{FeudalGain} & \multicolumn{2}{c}{STRAC-S} &\multicolumn{2}{c}{STRAC-M} &\multicolumn{2}{c}{HDC}\\
    \multicolumn{2}{c}{Task}&  Suc. & Rew. & Suc. & Rew. & Suc. & Rew. & Suc. & Rew.\\
    \hline
    \multicolumn{8}{c}{After 400 dialogues} \\
    \hline
    \multirow{3}{*}{ Env1 }  & CR &\textbf{99.8} & \textbf{\textcolor{blue}{14.2}}
 & 97.7 & 13.1
 &99.7 &14.0 &100.0 &14.1\\
    & SFR & 95.8 & 11.6
 & \textbf{98.2} & \textbf{\textcolor{blue}{12.3}}
 &99.2 &12.9 &97.6 &12.1\\
    & LAP & 96.1 & 11.5
 & \textbf{98.5} & \textbf{\textcolor{blue}{12.3}}
 &98.6 &12.2 &97.2 &11.8\\
    \hline
        \multirow{3}{*}{ Env2 }  & CR & \textbf{86.8} & \textbf{\textcolor{blue}{10.7}}
 & 65.5 & 5.0 &90.3 &10.2 &100.0 &14.1\\
    & SFR & \textbf{89.6} & \textbf{\textcolor{blue}{10.5}}
 & 69.8 & 4.4 &87.5 &9.0 &97.6 &12.4\\
    & LAP & \textbf{84.7} & \textbf{\textcolor{blue}{9.7}}
 & 56.9 & 1.6 &89.2 &9.1 &97.8 &11.7\\
    \hline
        \multirow{3}{*}{ Env3 }  & CR & \textbf{97.6} & \textbf{\textcolor{blue}{12.7}}
 & 97.2 & 12.5
 &97.3 &12.7 &95.2 &10.8\\
    & SFR & \textbf{91.4} & \textbf{\textcolor{blue}{9.1}} & 90.4 & 8.9
 &93.6 &10.5 &90.2 &8.9\\
    & LAP & 91.2 & 9.3 & \textbf{92.5} & \textbf{\textcolor{blue}{9.7}}
 &92.4 &9.6 &88.2 &8.4\\
    \hline
        \multirow{3}{*}{ Env4 }  & CR & \textbf{82.9} & \textbf{\textcolor{blue}{8.9}}
 & 71.0 & 5.2 &75.3 &6.6 &97.0 &11.1\\
    & SFR & \textbf{84.1} & \textbf{\textcolor{blue}{8.7}}
 & 72.7 & 5.0 &77.2 &6.4 &89.2 &8.2\\
    & LAP & \textbf{82.3} & \textbf{\textcolor{blue}{8.0}}
 & 65.9 & 3.1 &79.8 &6.9 &88.6 &8.4\\
    \hline
        \multirow{3}{*}{ Env5 }  & CR & 95.0 & \textbf{\textcolor{blue}{11.1}}
 & \textbf{95.3} & 10.6 &95.6 &10.8 &94.6 &9.2\\
    & SFR & \textbf{87.1} & \textbf{\textcolor{blue}{6.9}} & 80.6 & 4.5
 &88.8 &7.5 &87.6 &6.3\\
    & LAP & 87.6 & \textbf{\textcolor{blue}{6.6}} & \textbf{87.8} & 6.1
 &86.0 &5.6 &82.8 &4.6\\
    \hline
    
            \multirow{3}{*}{ Env6 }  & CR & \textbf{92.8} & \textbf{\textcolor{blue}{10.9}}
 & 91.9 & 10.3
 &90.7 &9.9 &91.2 &9.5\\
    & SFR & \textbf{79.4} & \textbf{\textcolor{blue}{5.6}}
 & 78.5 & 4.9 &83.8 &6.6 &80.2 &6.5\\
    & LAP & 81.7 & 6.1
 & \textbf{84.6} & \textbf{\textcolor{blue}{6.6}} &81.7 &5.7 &76.6 &5.6\\
    \hline
    
                \multirow{3}{*}{ Mean }  & CR & \textbf{92.5} & \textbf{\textcolor{blue}{11.4}}
 & 86.4 & 9.5 &91.5 &10.7 &96.3 &11.5\\
    & SFR & \textbf{87.9} & \textbf{\textcolor{blue}{8.7}}
 & 81.7 & 6.7 &88.3 &8.8 &90.4 &9.1\\
    & LAP & \textbf{87.3} & \textbf{\textcolor{blue}{8.5}}
 & 81.0 & 6.6 &88.0 &8.2 &88.5 &8.4\\
    \hline
\multicolumn{8}{c}{After 4000 dialogues} \\
    \hline
    \multirow{3}{*}{ Env1 }  & CR &\textbf{99.9} & \textbf{\textcolor{blue}{14.1}}
 &99.8 & \textbf{\textcolor{blue}{14.1}}
 &99.8 &14.1 &100.0 &14.1\\
    & SFR & \textbf{99.0} & \textbf{\textcolor{blue}{12.7}}
 & 98.7 & \textbf{\textcolor{blue}{12.7}}
 &98.5 &12.7 &97.6 &12.1\\
    & LAP & \textbf{97.6} & \textbf{\textcolor{blue}{12.0}}
 & \textbf{97.6} & \textbf{\textcolor{blue}{12.0}}
 &97.8 &12.0 &97.2 &11.8\\
    \hline
        \multirow{3}{*}{ Env2 }  & CR & \textbf{98.0} & \textbf{\textcolor{blue}{13.3}}
 & 97.9 & 13.1 &98.4 &13.1 &100.0 &14.1\\
    & SFR & \textbf{98.8} & \textbf{\textcolor{blue}{13.7}}
 & 95.6 & 12.1 &97.5 &13.0 &97.6 &12.4\\
    & LAP & \textbf{98.5} & \textbf{\textcolor{blue}{13.3}}
 & 92.6 & 11.6 &98.0 &12.8 &97.8 &11.7\\
    \hline
        \multirow{3}{*}{ Env3 }  & CR & \textbf{98.6} & \textbf{\textcolor{blue}{13.0}}
 & 98.1 & \textbf{\textcolor{blue}{13.0}}
 &97.9 &12.9 &95.2 &10.8\\
    & SFR & \textbf{95.1} & 10.4 & 91.9 & \textbf{\textcolor{blue}{10.5}}
 &93.0 &10.6 &90.2 &8.9\\
    & LAP & \textbf{91.2} & 9.5 & 90.7 & \textbf{\textcolor{blue}{9.7}}
 &92.1 &9.9 &88.2 &8.4\\
    \hline
        \multirow{3}{*}{ Env4 }  & CR & \textbf{98.0} & \textbf{\textcolor{blue}{12.5}}
 & 92.9 & 11.5 &91.3 &10.8 &97.0 &11.1\\
    & SFR & \textbf{93.1} & \textbf{\textcolor{blue}{11.2}}
 & 90.2 & 10.7 &89.2 &10.5 &89.2 &8.2\\
    & LAP & \textbf{91.8} & \textbf{\textcolor{blue}{11.1}}
 & 86.3 & 9.2 &89.7 &10.4 &83.6 &8.4\\
    \hline
        \multirow{3}{*}{ Env5 }  & CR & \textbf{97.6} & \textbf{\textcolor{blue}{11.9}}
 & 97.1 & 11.8 &96.5 &11.7 &94.6 &9.2\\
    & SFR & 88.4 & 7.1 &\textbf{ 89.6} & \textbf{\textcolor{blue}{8.4}}
 &90.1 &8.4 &87.6 &6.3\\
    & LAP & 87.6 & 6.6 & \textbf{88.2} & \textbf{\textcolor{blue}{6.9}}
 &88.5 &7.0 &82.8 &4.6\\
    \hline
    
            \multirow{3}{*}{ Env6 }  & CR & \textbf{94.0} & \textbf{\textcolor{blue}{11.0}}
 & 92.5 & \textbf{\textcolor{blue}{11.0}}
 &91.5 &10.7 &91.2 &9.5\\
    & SFR & \textbf{88.2} & \textbf{\textcolor{blue}{7.9}}
 & 81.6 & 7.0 &84.5 &7.3 &80.2 &6.5\\
    & LAP & \textbf{83.4} & \textbf{\textcolor{blue}{6.8}}
 & 83.3 & 6.7 &83.2 &6.7 &76.6 &5.6\\
    \hline
    
                \multirow{3}{*}{ Mean }  & CR & \textbf{97.7} & \textbf{\textcolor{blue}{12.6}}
 & 96.4 & 12.4 &95.9 &12.2 &96.3 &11.5\\
    & SFR & \textbf{93.8} & \textbf{\textcolor{blue}{10.5}}
 & 91.3 & 10.2 &92.1 &10.4 &90.4 &9.1\\
    & LAP & \textbf{91.7} & \textbf{\textcolor{blue}{9.9}}
 & 89.8 & 9.4 &91.6 &9.8 &88.5 &8.4\\
    \hline

\end{tabular}
}
\end{center}
    \caption{Success rate and average reward (using only $r_e$) for our proposed approach FeudalGain against STRAC-S. Best performance is marked in bold. Algorithms were tested in the Cambridge Restaurant (CR), San-Francisco Restaurant (SFR) and Laptops (LAP) domain. We included STRAC-M and a hand-coded policy (HDC) as supplemental comparison. Note that STRAC-M is trained in three domains and therefore utilized three times the amount of data compared to STRAC-S and FeudalGain. Results of STRAC were taken from \cite{strac}.}
\label{tab:sotatable}
\end{table}

\begin{center}
\begin{figure}[t]
\includegraphics[trim=0cm 0cm 0.0cm 1.0cm, width=0.51\textwidth]{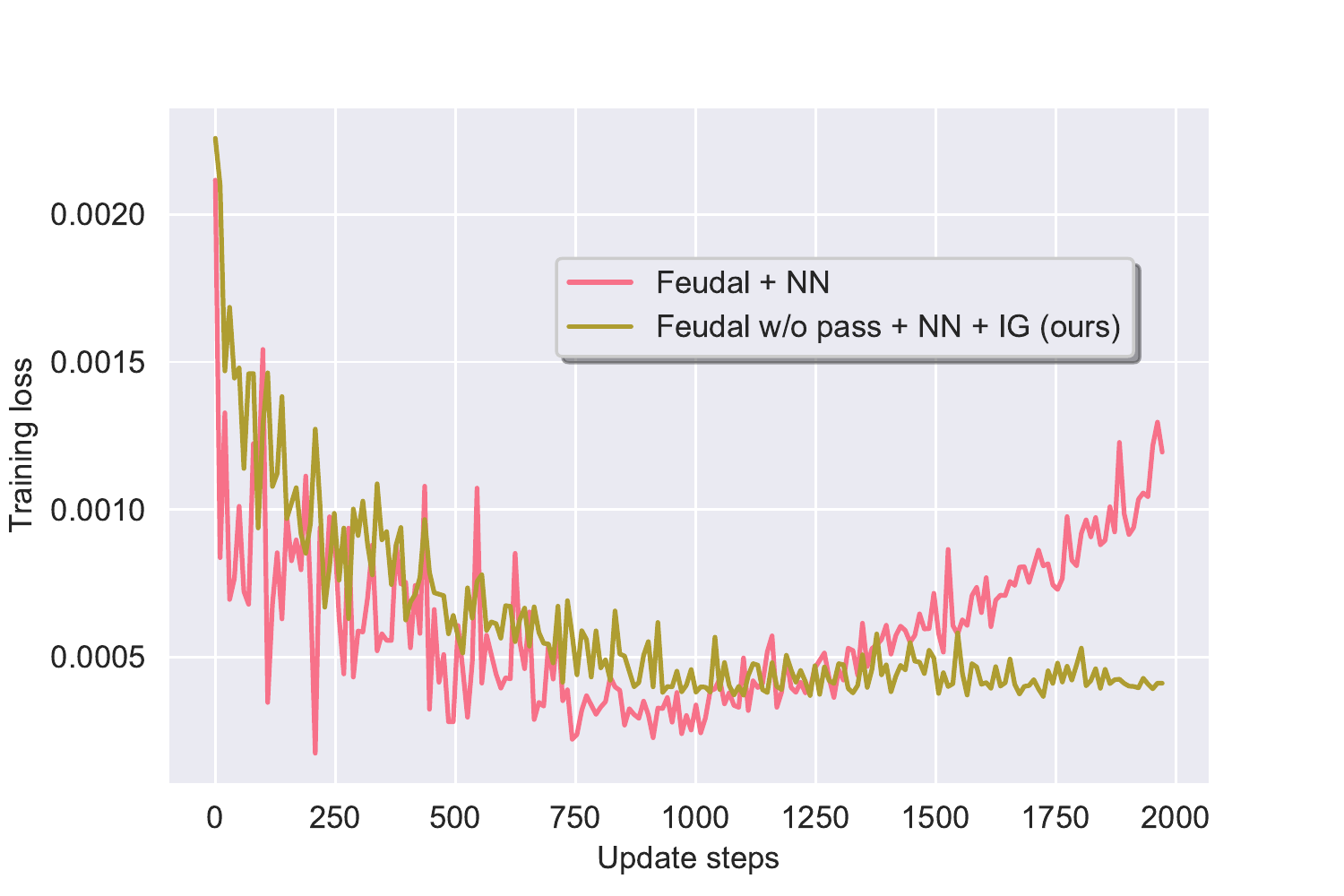}
\caption{Training loss for Feudal \cite{feudal2} using noisy networks (NN) and our proposed information gain reward (IG). We approximated the loss in every update step by taking 512 samples from the replay buffer.}
\label{lossfigure}
\end{figure}
\end{center}

%
%

\vspace{-1.2cm} 
\subsection{Ablation Study} \label{ablationstudy}

We conduct an ablation study for FeudalGain to investigate the difference in stability and convergence speed due to our proposed changes. We conduct experiments in environment 3 that exhibits a semantic error rate of $15 \%$. We chose this environment as it is close to human experiment characteristics \cite{env3proof}. Figure \ref{fig:feudalresult} depicts our findings. 

We first ablate the \emph{pass} action for $\pi_i$ by only updating $\pi_i$ with tuples $(b_t, a_t, r_t, b_{t+1})$ where $a_t \neq \text{\emph{pass}}$ to empirically verify that it is needed in order to back-propagate the final reward to actions taken by $\pi_i$. The algorithm is not capable of learning without \emph{pass} when only the extrinsic reward is used.

While usage of noisy networks can stabilise the learning for Feudal, the true benefit comes from the addition of information gain that results in fast and smooth convergence after as few as 500 dialogues. The dependence on the seed almost vanishes when introducing information gain, hence more stable learning and robustness against randomness in the initialisation is achieved. The usage of a single policy $\pi_{mg}$ in addition to information gain only led to a small performance difference in the first 200 dialogues. We hence omitted it in Figure \ref{fig:feudalresult} for better readability.

We further test the robustness of FeudalGain for increasing amount of semantic error rates by comparing it against Feudal+NN that does not use information gain. Results are depicted in Figure \ref{fig:noisetest}, for policies trained for 200 and 4000 dialogues, averaged over 15 seeds. For 200 training dialogues, the difference between FeudalGain and the baseline is consistent across noise levels. For 4000 training dialogues, the baseline catches up a little but does not outperform FeudalGain on any of the noise levels.

Lastly, we want to empirically verify that the usage of extrinsic reward may provide misleading feedback for $\pi_i$, resulting in incorrect policy updates. Figure \ref{lossfigure} shows the training loss for policy $\pi_i$ using Feudal with noisy networks and information gain. Substituting $r_e$ for $r_i^\prime$ leads to quick and stable convergence of the algorithm with weaker oscillations.

\subsection{Human Evaluation}
In order to show that our results transfer from simulation to humans, we compare FeudalGain against Feudal with noisy networks (Feudal + NN) in a human trial, where users directly interact with the two policies.
We collected 400 dialogues using each policy. We took the policies after only 200 training dialogues that were closest to the average performance in environment 3. The reward was $11.7$ for Feudal + NN and $12.9$ for FeudalGain on the simulated user. We chose such a small number of training dialogues to examine the sample efficiency of FeudalGain. At the end of each interaction, we asked users if the dialogue was successful, whether the system asked for information when necessary (``AskIfNec'') and what the overall performance was (``Overall''). Table \ref{humaneval1} shows that superior performance of FeudalGain in terms of success and number of turns in simulation translates to real users. More interestingly, the rating if information was requested when necessary is much higher, which confirms that our intrinsic reward enables $\pi_i$ to learn guided information gathering. The overall rating (``Overall'') is correlated to ``AskIfNec'', showing how important guided information gathering is for the overall perception of the system. The reduced standard deviation shows the stability of our approach.

\begin{table}
\centering
\resizebox{\columnwidth}{!}{
\begin{tabular}{l|c|c|c|c}
 & Success & Turns & AskIfNec & Overall\\
 \hline
FeudalGain & $0.71/0.45^\ast$ & $6.5/3.4^\ast$ & $3.8/1.4^\ast$ & $3.7 / 1.5^\ast$\\
\hline
Feudal + NN & $0.43 / 0.5$ & $8.1/5.0$ & $3.0/1.5$ & $2.7/1.6$\\
\hline
\end{tabular}
}
\caption{Mean/standard deviation for success, number of turns, whether the system asked for information when necessary and overall performance according to human evaluation. $^\ast$We used the t-test to check statistical significance, where $p<0.05$.}
\label{humaneval1}
\end{table}

\section{Conclusion}

We proposed the use of intrinsic reward within the hierarchical Feudal Dialogue Management approach for the information seeking policy. Our new architecture gracefully deals with shortcomings such as artificial \emph{pass} actions and misleading reward signals that lead to sample inefficiency and instability. Our proposed reward encourages the policy to seek useful information from the user and puts more emphasis on the user's needs, which is an integral part of dialogue systems that aid the user in solving any kinds of tasks. We show in experiments with simulated users that incorporating our reward improves sample efficiency, stability, and quality of the resulting policy, and that our algorithm FeudalGain algorithm leads to state-of-the-art results for the PyDial benchmark. We confirm the results in a human trial where volunteers interacted with our policy. Our results warrant a more widespread use of intrinsic reward in task-oriented dialogue systems.

In future work, we like to scale our approach to multiple domains and learn the hierarchical structure automatically.

\section*{Acknowledgments}

This work is a part of DYMO project which has received funding from the European Research Council (ERC) provided under the Horizon 2020 research and innovation programme (Grant agreement No. STG2018 804636). N. Lubis, C. van Niekerk, M. Heck and S. Feng are funded by an Alexander von Humboldt Sofja Kovalevskaja Award endowed by the Federal Ministry of Education and Research. Computational resources were provided by Google Cloud.


\bibliographystyle{IEEEbib}
\bibliography{strings,refs}

\end{document}